\documentclass{article}
\usepackage{spconf,amsmath,graphicx}
\usepackage{algorithm}
\usepackage{algorithmic}
\usepackage{epsfig}
\usepackage{subfigure}
\usepackage{subfig}
\usepackage{amsmath}
\usepackage{amssymb}
\usepackage{graphicx}

\title{ENHANCING DATA-FREE ADVERSARIAL DISTILLATION WITH ACTIVATION REGULARIZATION AND VIRTUAL INTERPOLATION}
\name{Xiaoyang Qu, Jianzong Wang$^*$, Jing Xiao\thanks{Corresponding author: Jianzong Wang (e-mail: jzwang@188.com).}}
\address{Ping An Technology (Shenzhen) Co., Ltd.}
\begin{document}
%
\maketitle
\begin{abstract}
Knowledge distillation refers to a technique of transferring the knowledge from a large learned model or an ensemble of learned models to a small model. This method relies on access to the original training set, which might not always be available. A possible solution is a data-free adversarial distillation framework, which deploys a generative network to transfer the teacher model's knowledge to the student model. However, the data generation efficiency is low in the data-free adversarial distillation. We add an activation regularizer and a virtual interpolation method to improve the data generation efficiency. The activation regularizer enables the students to match the teacher's predictions close to activation boundaries and decision boundaries. The virtual interpolation method can generate virtual samples and labels in-between decision boundaries. Our experiments show that our approach surpasses state-of-the-art data-free distillation methods. The student model can achieve 95.42\% accuracy on CIFAR-10 and 77.05\% accuracy on CIFAR-100 without any original training data. Our model's accuracy is 13.8\% higher than the state-of-the-art data-free method on CIFAR-100.
\end{abstract}
\section{Introduction} \label{sec:intro}
In recent years, deep neural networks show remarkable performances in various machine learning applications. However, deep networks have improved in performance at the cost of larger model size and higher computational overhead, which hinders the deployment on mobile devices or Internet-of-Things devices. This problem can be alleviated by various model compression methods, such as model pruning\cite{deepcompression},  quantization\cite{quantization_xnornet}, and knowledge distillation\cite{TSC_First,KD_FEATURE_Fitnets,KD_FEATURE_FT,KD_FEATURE_AT,KD_FEATURE_Similarity}.

These model compression paradigms are commonly relying on accessing original data.  After pruning or quantizing to a squeezed model, it requires original data to retrain the compressed model. The modern knowledge distillation trains the student model using original data and corresponding labels produced by a powerful teacher model. However, most of original training datasets are unavailable in real-world scenarios. Besides, large datasets make the transmission prohibitively expensive. Moreover, privacy and security concerns prohibit the distribution of the training data. Thus, it is necessary and important to design a data-free model compression method without accessing original data.


Data-free knowledge distillation\cite{DF_DeepInversion,DF_DreamDistillation,DF_noObservableData,DF_Quantization,DF_ModelAttack,DF_Huawei,DF_Meta,DF_ZeroShot_BeliefMatching,DF_AliKD} can utilize these pre-trained models to accomplish model compression without accessing original data. Data-free knowledge distillation using metadata or a generative network to construct training distribution, which could be extracted from a record of the teacher's training activations. There are two optional ways to implement data-free methods. One way is to use similar data or metadata to approximate the training distribution. However, the metadata and similar data are unavailable in most real-world applications. The other way is to use a generator to construct the training data distribution to help transfer the knowledge from a pre-trained model to a small model. But it is still challenging to reconstruct an accuracy data distribution by extracting from the teacher model's activations. 

This work focuses on the data-free adversarial distillation framework, which deploys a generator to help transfer the knowledge from a teacher model to a student model. The conventional adversarial distillation aims to create abnormal samples to hinder the student model from mimicking the teacher model's behavior. Although these abnormal samples can generate samples in the low-density region where the decision boundary lies, the data generation efficiency is low. Thus, we enhance data generation efficiency by integrating an activation regularizer, enabling students to match the teacher's predictions close to activation boundaries and decision boundaries. Besides, we apply a virtual interpolation to construct virtual examples in-between boundary decisions.

The main contributions are shown as follows:  (1) An activation regularizer is integrated into data-free adversarial distillation to enhance data generation efficiency. (2) A virtual interpolation method is deployed to improve adversarial distillation by generating virtual samples and labels in-between decision boundaries. (3) Our experiments show that we obtain a performance 13.8\% higher than the state-of-the-art data-free method on CIFAR-100. Moreover, our approach can achieve comparable results with data-driven methods.

\begin{figure*}[!htb]
  \centering
    \subfigure[Data-free adversarial framework ]{\includegraphics[width=0.45\textwidth]{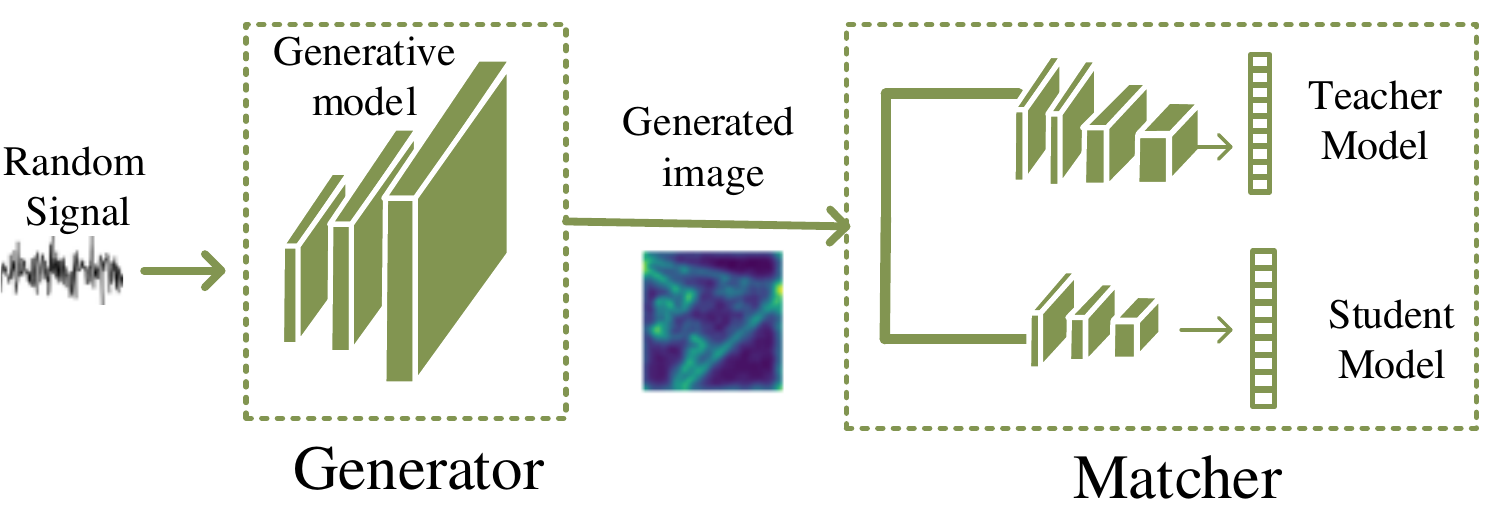}}   
     \subfigure[The visualization of generated images ]{\includegraphics[width=0.42\textwidth]{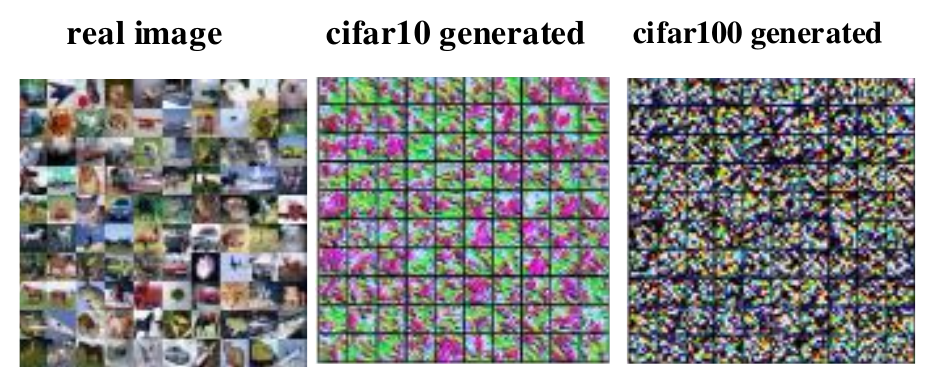}}   
    \subfigure[]{\includegraphics[width=0.2\textwidth]{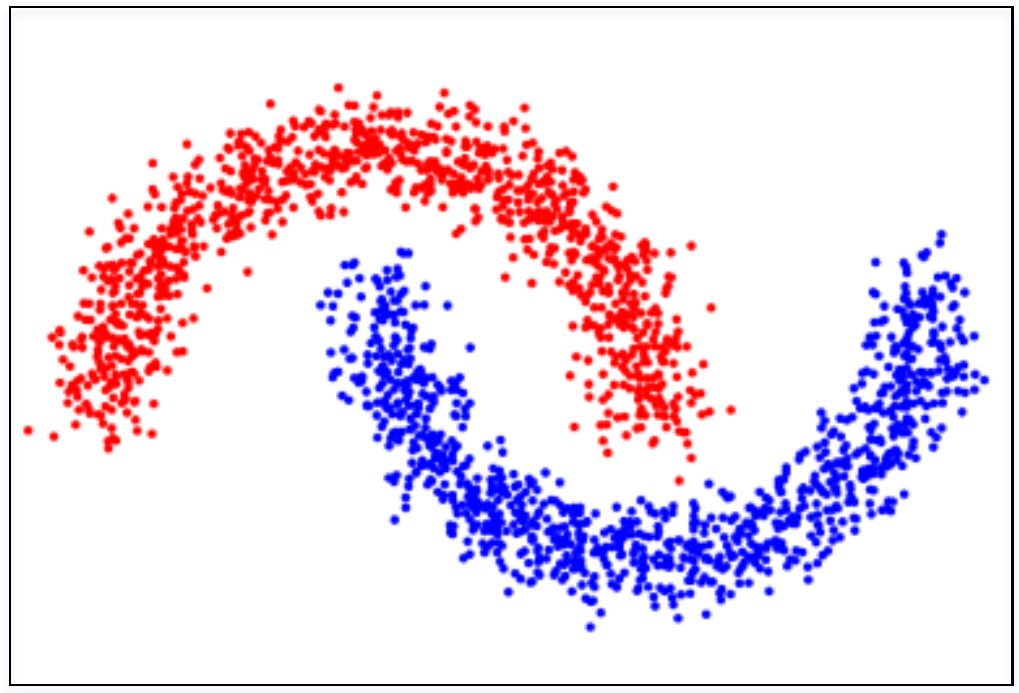}}  
    \subfigure[]{\includegraphics[width=0.2\textwidth]{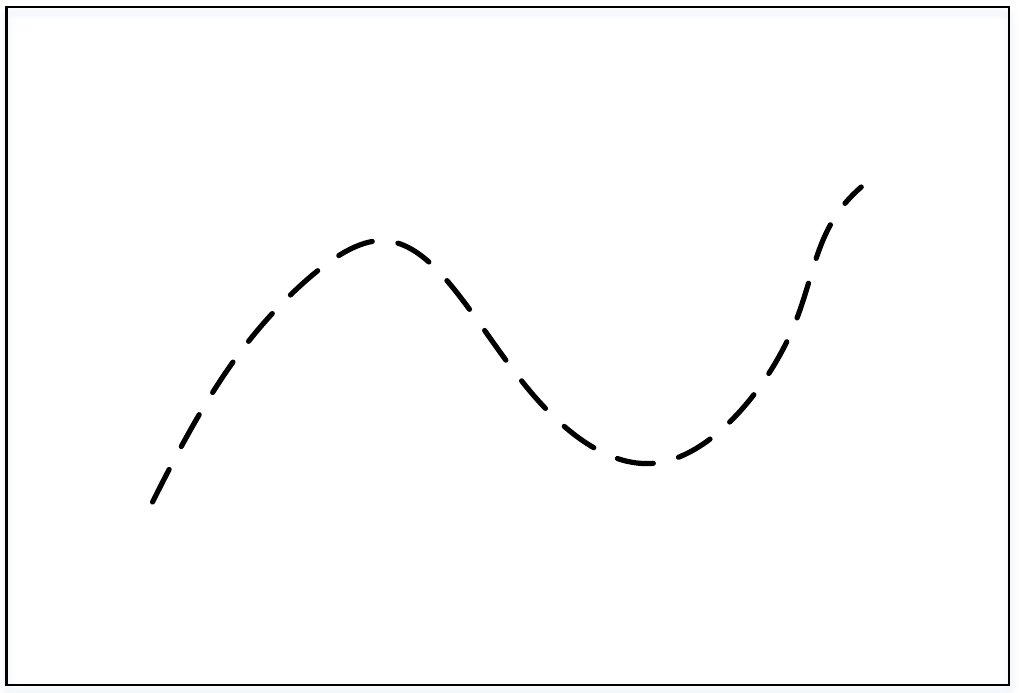}}  
    \subfigure[]{\includegraphics[width=0.2\textwidth]{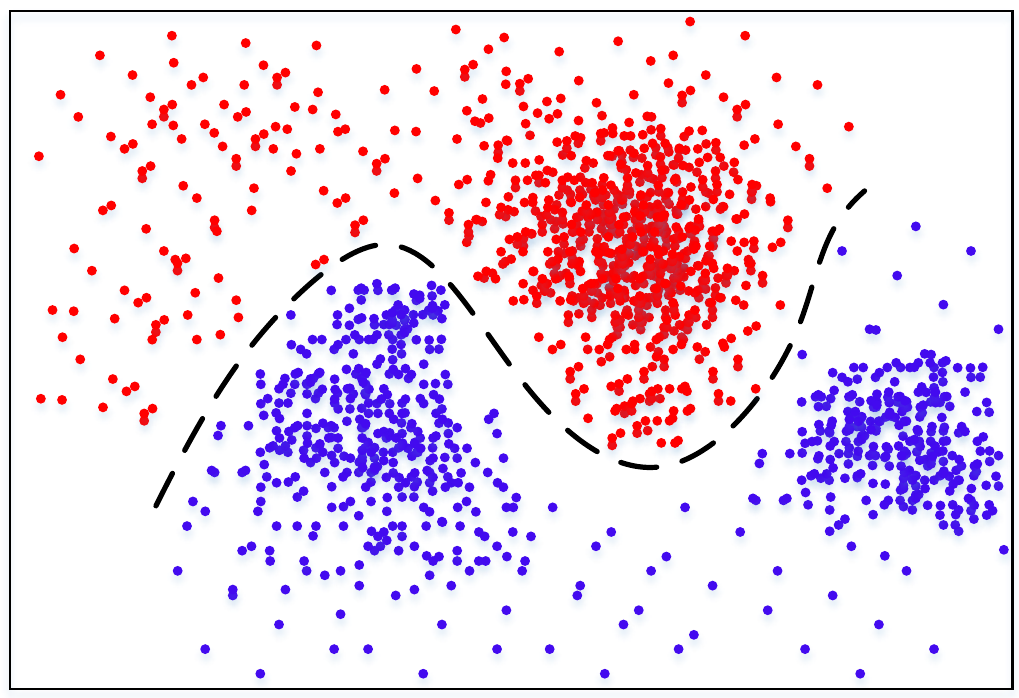}}  
    \subfigure[]{\includegraphics[width=0.2\textwidth]{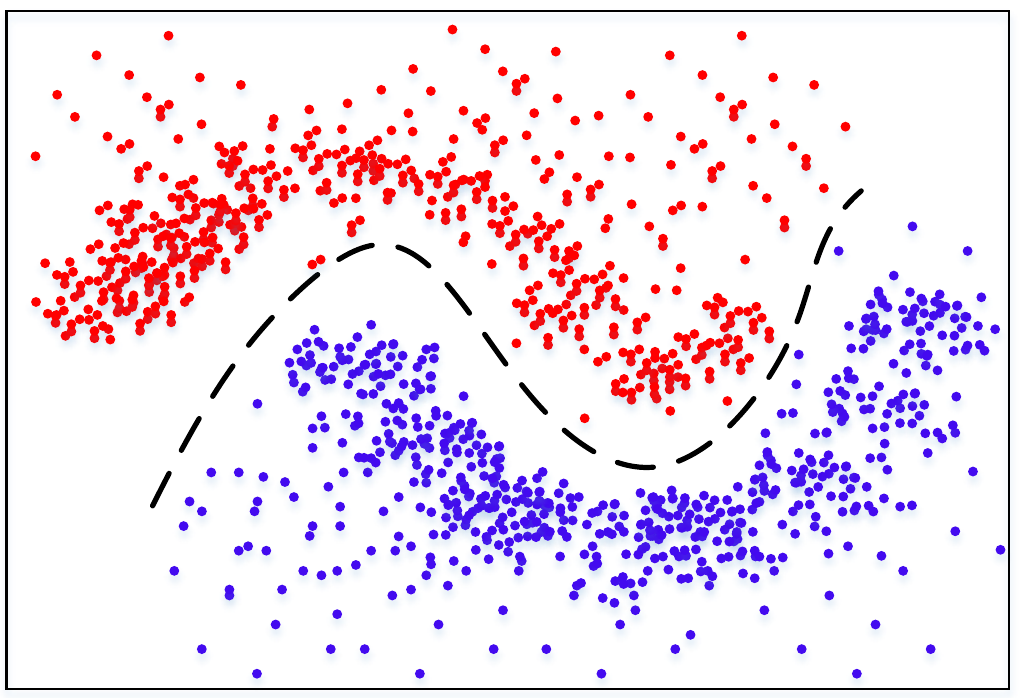}} 
    \vskip -0.15in 
    \caption{(c) The two-moon dataset. (d) The decision boundary of the teacher model. (e) Points generated by conventional data-free adversarial distillation. (f) Points generated by our methods }
    \label{fig:mainidea}
    \vskip -0.1in
\end{figure*}

\section{Preliminaries and Motivation}
\subsection{Data-Free Adversarial distillation}

Typical data-free adversarial framework \cite{DF_AliKD}\cite{DF_ZeroShot_BeliefMatching} deploys a generator to transfer the knowledge of the pre-trained teacher model to a student model without access to original data. As shown in Figure \ref{fig:mainidea}(a),  data-free adversarial framework deploys a \textit{generator} to transfer the knowledge of a pre-trained \textit{teacher} model to a \textit{student} model without any original data. Here, the teacher model and the student model jointly act as a \textit{matcher}. Adversarial learning means \textit{generator} and \textit{matcher} compete with each other to achieve their individual goals. The \textit{matcher} aims to minimize the model discrepancy, but the \textit{generator} seeks to maximize the model discrepancy. 

Why a student network trained on samples generated by adversarial distillation can give good performance? While naive adversarial training\cite{GAN_START} aims to generate realistic and indistinguishable samples to fool the discriminator, the adversarial distillation generates abnormal examples to hinder teacher-student capacity matching. In this way, the adversarial distillation can generated as many samples as possible to cover the input space. However, the data generation efficiency is low in the conventional adversarial distillation.




\subsection{Motivation}
Let us review the cluster assumption\cite{cluster_assumption}: \textit{if two samples belong to the same cluster in the input distribution, then they are likely to belong to the same class}. The cluster assumption is similar to the low-density separation assumption\cite{low_density_assumption}: \textit{the decision boundary should lie in the low-density regions}. We conduct a toy experiment on the two-moon dataset, as shown in Figure\ref{fig:mainidea}(c). We trained a teacher model on this two-moon dataset. We can find the teacher model moves the decision boundary to low-density regions, as shown in Figure\ref{fig:mainidea}(d). As a classification task highly depends on decision boundaries among classes, the generator in conventional adversarial distillation can generate as many samples as possible to cover the input space. Besides, adversarial distillation generated points towards a region of low density. Thus this is a good direction to move the decision. As shown in Figure \ref{fig:mainidea}(e), the framework generates samples covering enough input space and towards a region of low density.




However, the conventional data-free adversarial distillation is limited to the low data generation efficiency. It may generate abnormal samples, which result in deactivated neurons. Figure \ref{fig:mainidea}(b) shows the visualization of  images generated by the conventional data-free adversarial distillation\cite{DF_AliKD}. we can find the generated images are not realistic images in intuition. Namely, we can easily discriminate between the real and generated images. The data generation efficiency is worse in a larger region. Thus \textit{the main motivation of this work to improve the data generation efficiency for adversarial distillation}. Namely, we aim to discover a method, which needs fewer samples generated to achieve good performance.

\section{Proposed Method} \label{sec:Matching}
\subsection{Activation Regularization}


As adversarial distillation can generate samples towards the low-density region where decision boundary lies, our method developed from the conventional data-free adversarial distillation. The adversarial goal can be formulated as Equation 1 using the student network $S$ with parameters $\theta$ and the generation network $G$ with parameters $\phi$.
\begin{equation}
\left\{\begin{matrix}
S^{*}=\underset{S}{min}\underset{\theta}{L}(T(G(z)),S(G(z);\theta))  \\
G^{*}=\underset{G}{max}\underset{\phi}{L}(T(G(z;\phi)),S(G(z;\phi)))
\end{matrix}\right.
\end{equation}
where $T$ means the the teacher model, $z$ means random signal, $L(\cdot)$ means the loss function. 

As the transfer of activation boundaries can help generate the high-region samples, we add two-level regularization into the framework to reduce unnecessary abnormal samples generated. The logits-level activation regularization $R_{l}$ is shown as follows.
\begin{equation}
    R_{l}=\frac{1}{n}\sum_i^N f_{CE}(l^i_T,y_i)
\end{equation}
where $f_{CE}(\cdot)$ means the cross entropy function, $l_T^i$ is the predicted logits of the teacher model, and $y_i$ means the pseudo label produced by the teacher model. The representation-level activation regularizer $R_{m}$ is formulated as 
\begin{equation}
    R_{m}=-\frac{1}{n}\sum_i\left \| f_T^i \right \|_1
\end{equation}
where $\left \| \cdot \right \|$ is the conventional $l_1$ norm. $f_T^i$ means features of $x^i$ generated by the teacher model.

The activation regularization can transfer the activation boundaries, which refers to a separating hyperplane that determines whether the neuron is activated or deactivated. Some works\cite{ActivationBoundary1,ActivationBoundary2} provide the fact that the decision boundaries is composed of activation boundaries. The data generation efficiency will be significantly improved in a larger region of input space. As shown in Figure 2(f), our adversarial distillation with activation regularization has similar input spaces.



\subsection{Virtual  Interpolation} \label{sec:Optimization}
To further enhance the data generation efficient, we use a local linear regularizer to construct virtual examples in-between boundary decisions. We expect the decision boundary transition linearly from class to class. For linear interpolation, Mixup\cite{Regula_Mixup,Regula_AdaMixup,Regula_ManifoldMixup} is a proper choice. Let us reviews the fundamental conception of Mixup. Mixup makes random interpolation between two examples, along with identical interpolation between corresponding labels. In intuition, Mixup constructs virtual training examples in-between decision boundaries. The virtual samples $x_v$ and corresponding virtual label $y_v$ are calculated as  
\begin{equation}
\begin{cases}
x_v= \lambda x_i+(1-\lambda)x_j \\
y_v= \lambda y_i+(1-\lambda)x_j 
\end{cases}
\end{equation}
Here, $x_i,x_j $ are raw input vectors, and $y_i,y_j$ are one-hot label encodings. And $ \lambda \in [0,1]$ is a coefficient. In intuition, Mixup is a local linear regulation. The connection between the virtual sample $x_v$ and the virtual label $y_v$ is $y_v=f(x_v)$. So we cat get 
\begin{equation}\label{equ:mixup_origin}
    \lambda y_i + (1-\lambda )y_j = f(\lambda x_i + (1-\lambda ) x_j)
\end{equation}
if we use $f(x_i)$ to replace $y_i$, and $f(x_j)$ to replace $y_j$, Equation \ref{equ:mixup_origin} can be converted into Equation \ref{equ:mixup_function}.
\begin{equation}\label{equ:mixup_function}
     \lambda f(x_i) + (1-\lambda )f(x_j) = f(\lambda x_i + (1-\lambda ) x_j)
\end{equation}

Assuming $\lambda, x_i, x_j$ are variables, a feasible solution to Equation \ref{equ:mixup_function} is any linear function $f$ worked in this Equation \ref{equ:mixup_function}. Thus, Mixup plays a role in interpolating a local linear regularizer. The optimization goal of the regularizer can be formulated as follows. \begin{equation}\label{equ:mixup_loss1}
\underset{f}{min}\:\underset{x_i,x_j}{E}\:L(\underbrace{f(\lambda x_i+(1-\lambda)x_j)},\underbrace{\lambda f(x_i)+(1-\lambda)f(x_j)})
\end{equation}
here, $L(\cdot,\cdot)$ means a loss function, such as mean square error function.$\underset{x_i,x_j}{E}(\cdot)$ means the expected value with respect to $x_i,x_j$. 

However, there is no label information in our framework. Thus we use the teacher model $f_T$ to produce the pseudo labels for student model $f_S$. Equation \ref{equ:mixup_loss1} can be transformed into Equation \ref{equ:mixup_loss2}
\begin{equation}\label{equ:mixup_loss2}
\underset{f_S}{min}\:\underset{x_i,x_j}{E}\:L(\underbrace{f_S(\lambda x_i+(1-\lambda)x_j)},\underbrace{\lambda f_T(x_i)+(1-\lambda)f_T(x_j)})
\end{equation}

As there are no realistic input vectors in our framework. We use a generator $f_G$ to map two random signals $z_i$ and $z_j$ to two pseudo image $f_G(z_i)$ and $f_G(z_i)$. if we use $f_G(z_i)$ to replace $x_i$, and $f_G(z_j)$ to replace $y_j$, then Equation \ref{equ:mixup_loss2} can be converted into Equation \ref{equ:mixup_loss3}.
\begin{equation}\label{equ:mixup_loss3}
\begin{matrix}
\underset{f_S, f_T}{min}\: \underset{z_i,z_j}{E}\: & L(f_S(\lambda f_G(z_i)+(1-\lambda)f_G(z_j)), \\
& \lambda f_T(f_G(z_i))+(1-\lambda)f_T(f_G(z_j)))
\end{matrix}
\end{equation}


Why do we incorporate the Mixup into our framework? Firstly, Mixup can construct virtual samples and virtual labels to preserve sample diversity. Secondly, Mixup makes the decision boundary more smooth by encouraging the model to behave linearly between training examples. Thus, decision boundaries in our framework can transition linearly from class to class. 

\begin{table*}[htbp]
\centering
\caption{The comparison on MNIST, CIFAR-10, and CIFAR-100}
\label{tab:overall}
\begin{tabular}{cc|cc|cccc}
\hline
\textbf{Model} & 
\textbf{Require Data} &
\textbf{FLOPs} & 
\textbf{\begin{tabular}[c]{@{}l@{}}Acc(\%)\\ (MNIST)\end{tabular}} & 
\textbf{FLOPs} & 
\textbf{Pars} & 
\textbf{\begin{tabular}[c]{@{}l@{}}Acc(\%)\\ (cifar10)\end{tabular}} & 
\textbf{\begin{tabular}[c]{@{}l@{}}Acc(\%)\\ (cifar100)\end{tabular}} \\ \hline
Teacher & original data & 433K  & 98.89 & 1.16G & 21M & 97.20 & 81.13 \\ 
\hline
KD\cite{TSC_Hinton} & original data & 139K  & 98.67 & 557M & 11M   & 95.55 & 77.61 \\ 
FITNET\cite{KD_FEATURE_Fitnets} & original data & 139K & - & 557M & 11M  & 96.24 & 80.49 \\ 
AT\cite{KD_FEATURE_AT} & original data & 139K & 98.42  & 557M & 11M  & 96.30 & 78.54   \\ 
FT\cite{KD_FEATURE_FT} & original data & 139K  & 97.84 &  557M & 11M  & 95.29 & 76.93  \\ \hline
DFL\cite{DF_Huawei} & no data & 139K & 98.20& 557M & 11M   & 88.5 & 61.4   \\ 
DFAL\cite{DF_AliKD} & no data & 139K & 98.30 & 557M & 11M   & 93.5 & 67.7   \\
\textbf{Proposed} & no data & 139K   & \textbf{98.41} & 557M & 11M  & \textbf{95.42} & \textbf{77.05}  \\ \hline
\end{tabular}
\vskip -0.2in
\end{table*}



\section{EXPERIMENTS} \label{sec:experiment}
\subsection{Setup}
Our method is evaluated on 3 different datasets. The baselines include three teacher models, four data-driven knowledge distillation methods, two data-free knowledge distillation methods. For \textbf{Teacher}, there are a given pre-trained LeNet-5 on MNIST dataset and a given pre-trained ResNet-34 on CIFAR-10 and CIFAR-100 datasets. For \textbf{KD}\cite{TSC_Hinton}, the student model is trained to match the soften target of the teacher model. For \textbf{FitNet}\cite{KD_FEATURE_Fitnets},  the student model is trained to match the logit and intermediate representation of the teacher model. For \textbf{AT}\cite{KD_FEATURE_AT}, the student model is trained to match the attention maps of the teacher model. For \textbf{FT}\cite{KD_FEATURE_FT}, the student model is trained to match the factors of the teacher model. For \textbf{DFL}\cite{DF_Huawei}, the student model is trained using Data-Free Learning(DFL) without accessing original data. For \textbf{DFAL}\cite{DF_AliKD}, the student model is trained using Data-Free Adversarial Learning(DFAL) without accessing original data.


\subsection{Experiment Results}
\textbf{Overall Evaluation}. Table \ref{tab:overall} shows the performance of different knowledge distillation methods. The baseline includes two kinds: data-driven knowledge distillation methods and data-free knowledge distillation methods. For MNIST, LeNet-5 and LeNet-5-HALF are deployed as the teacher model and the student model, respectively. For CIFAR-10 and CIFAR-100, we use ResNet-34 as the teacher model and ResNet18 as the student model. The teacher model is 97.2\% accuracy on CIFAR-10 and 81.13\% on CIFAR-100. On each dataset, our method goes beyond all data-free methods and partial data-driven methods. On MNIST, our method gets a small improvement overall data-free method, because the accuracy of the MNIST dataset is so high that there is no enough room for improvement. On CIFAR-10, our method gets 2\% improvement over DFAL and 7.8\% improvement over DFL. However, on CIFAR-100, our method gets 13.8\% accuracy improvement over the state-of-the-art data-free method DFAL. 

Our method achieved comparable performance to data-driven approaches, such as KD, FitNet, AT,  and FT.  On the MNIST dataset, the accuracy difference between our method and other data-free methods is no more than 1\%. On CIFAR-10, the situations are similar. However, on CIFAR-100, the accuracy differences between our method and other data-driven methods is no more than 3.44\%.



\begin{figure}[!htb]
  \centering
    \subfigure[The comparison between data-driven methods and our method]{\includegraphics[width=0.22\textwidth]{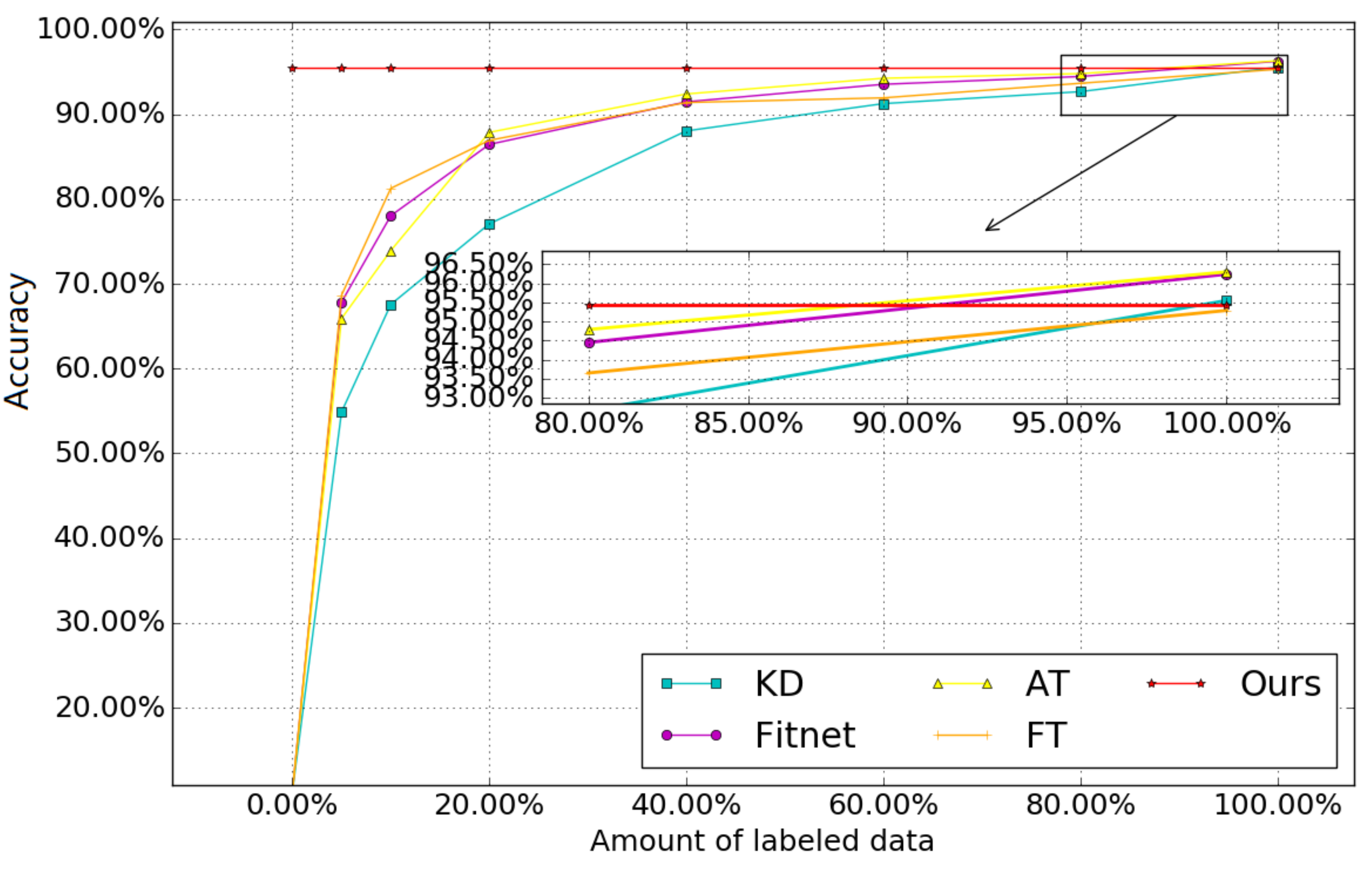}}  
    \subfigure[The comparison of data-free methods and our method]{\includegraphics[width=0.22\textwidth]{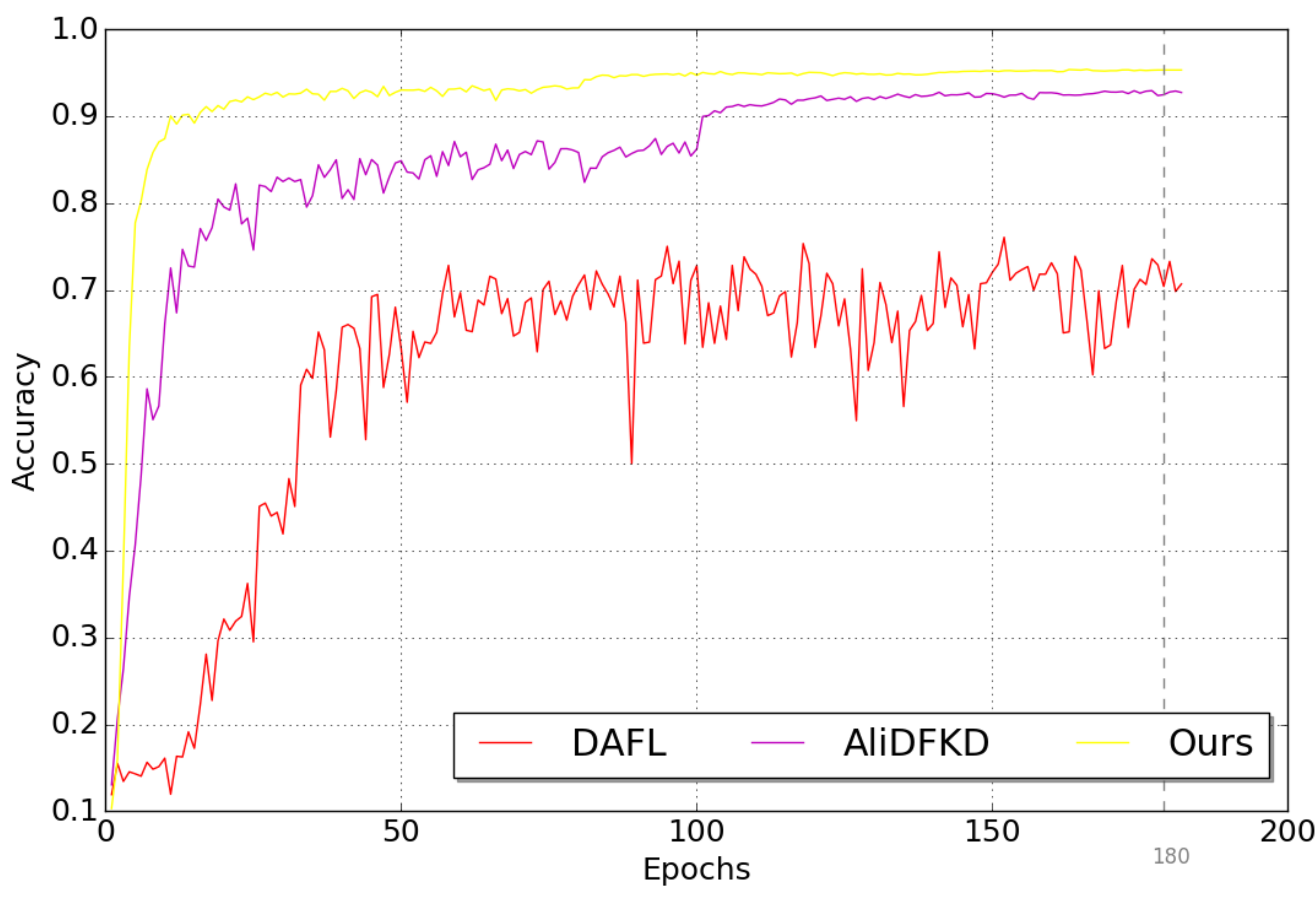}}
    \subfigure[Evaluation of various components on CIFAR-10]{\includegraphics[width=0.22\textwidth]{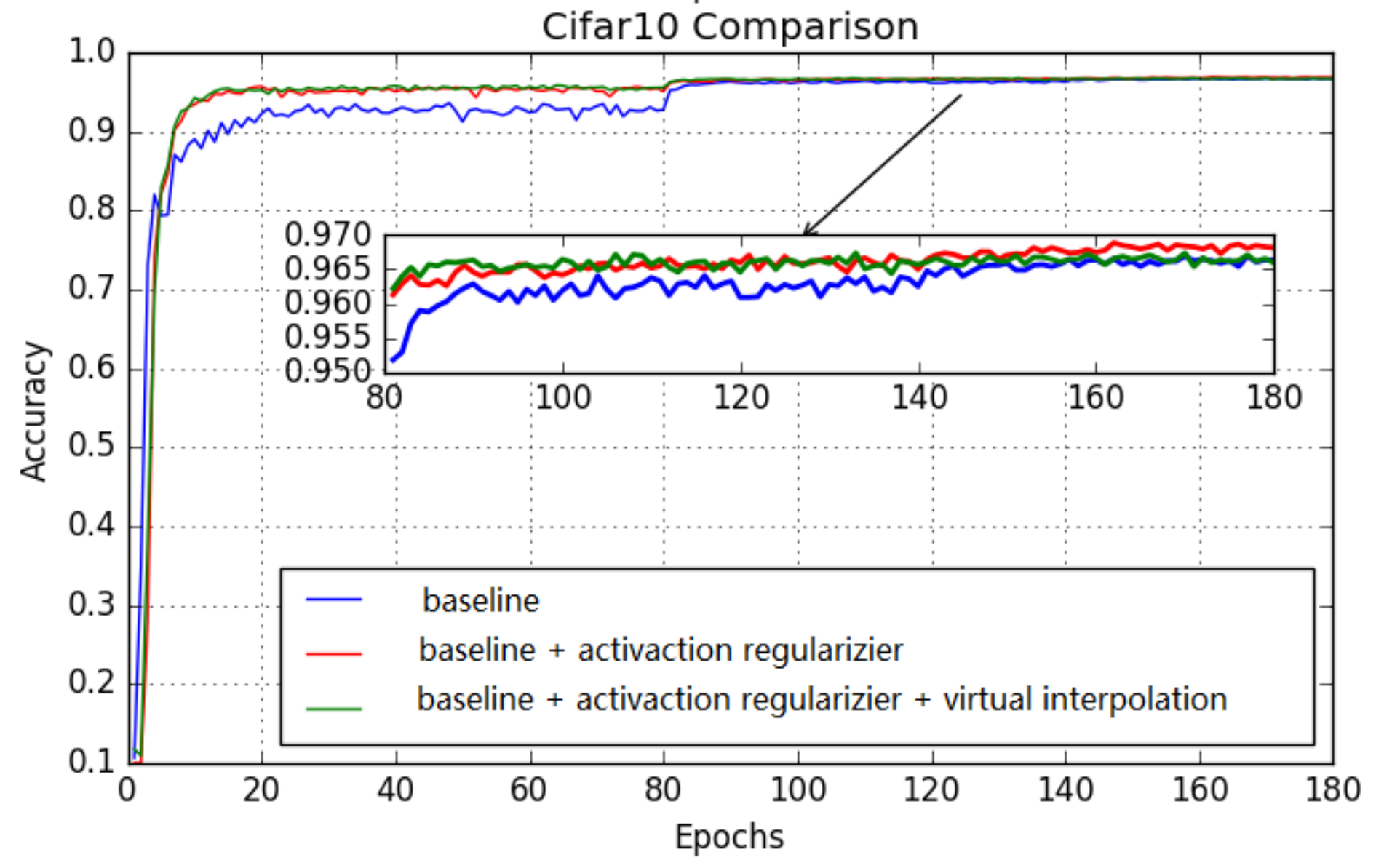}}  
    \subfigure[Evaluation of various components on CIFAR-100]{\includegraphics[width=0.22\textwidth]{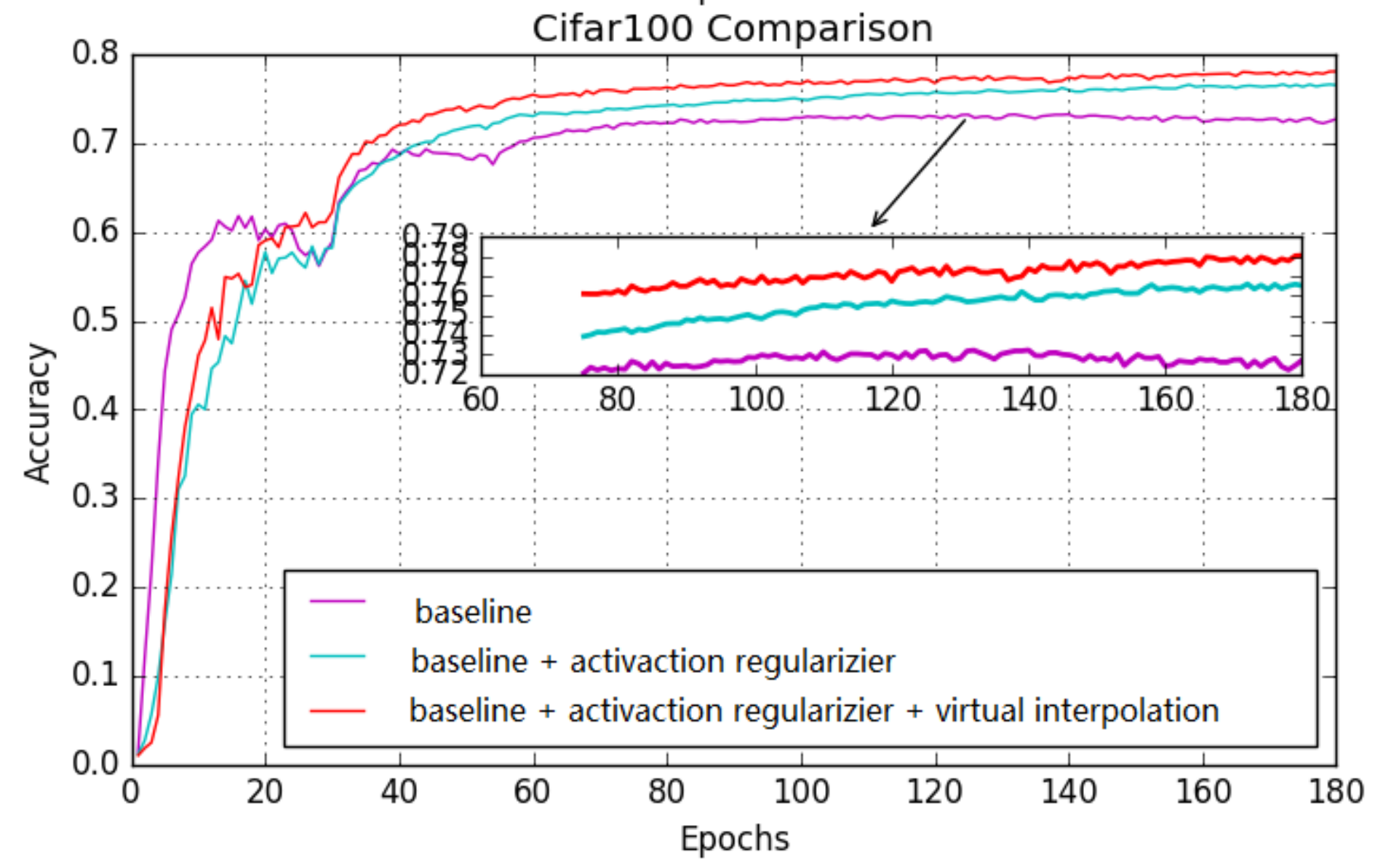}}
    \vskip -0.1in
    \caption{Various evaluation results of our method}
    \label{fig:SixCurve}
    \vskip -0.3in
\end{figure}

\textbf{Data Generation Efficient}. Figure \ref{fig:SixCurve}(a) shows the accuracy comparison between data-driven methods and our method, along with the increase of labeled data mount. We found our method performance is relatively constant across all amounts of labeled data. In the case of low labeled data amount, data-driven methods have very low accuracy, but our approach can reach 95.42\%. Figure \ref{fig:SixCurve}(b) shows the accuracy comparison between data-free methods and our method over time., DFAL and our method converge very quickly, but DFL converges very slow. In the $180_{th}$ epoch, DFL has not converged yet. In our experiment, DFL will converge in about $2000_{th}$ epochs. Figure \ref{fig:SixCurve}(c) and Figure \ref{fig:SixCurve}(d) presents the accuracy comparison of our framework with and without various components on CIFAR-10 and CIFAR-100. Using the regularizer and virtual interpolation, the convergence speed is faster, and the final convergence accuracy is higher.


\section{CONCLUSIONS}\label{sec:conclusion}
To improve data generation efficient, we incorporate two regularizers into adversarial distillation framework to let the student model trained in the activation boundaries and the decision boundaries. The experimental results demonstrate that our method can outperform all data-free methods. Notably, on CIFAR-100, our method gets 13.8 \% accuracy improvement over the state-of-the-art data-free method. Moreover, our method can achieve comparable results with data-driven approaches. The gap between our method and other data-driven methods is no more than 1\%.

\section{Acknowledgment}
This paper is supported by National Key Research and Development Program of China under grant No. 2017YFB1401202, No. 2018YFB1003500, and No. 2018YFB0204400. Corresponding author is Jianzong Wang from Ping An Technology (Shenzhen) Co., Ltd.

\bibliographystyle{IEEEtran}

\bibliography{refs/KD, refs/NAS, refs/SSL, refs/KDGAN, refs/DataFreeKD, refs/GAN, refs/introduction, refs/regularization}

\end{document}